%% file: main.tex
\documentclass[10pt,twocolumn,letterpaper]{article}

\usepackage{cvpr}              
\input{preamble}
\definecolor{cvprblue}{rgb}{0.21,0.49,0.74}
\usepackage[pagebackref,breaklinks,colorlinks,allcolors=cvprblue]{hyperref}

\title{Efficient Complex-Valued Vision Transformers for MRI Classification Directly from k-Space}

\author{
    Moritz Rempe\textsuperscript{1,2,4}, 
    Lukas T. Rotkopf\textsuperscript{1,4,7}, 
    Marco Schlimbach\textsuperscript{4},
    Helmut Becker\textsuperscript{1},
    Fabian Hörst\textsuperscript{1}, \\
    Johannes Haubold\textsuperscript{1,3,8}, 
    Philipp Dammann\textsuperscript{9},
    Kevin Kröninger\textsuperscript{4}, and 
    Jens Kleesiek\textsuperscript{1,2,3,4,5,6}
    \vspace{2mm} \\ 
    \small \itshape 
    \textsuperscript{1}Institute for AI in Medicine (IKIM), University Hospital Essen, Girardetstraße 2, 45131 Essen, Germany\\
    \small \itshape 
    \textsuperscript{2}Cancer Research Center Cologne Essen (CCCE), University Medicine Essen, Hufelandstraße 55, 45147 Essen, Germany\\
    \small \itshape 
    \textsuperscript{3}RACOON Study Group, Site Essen, Essen Germany\\
    \small \itshape 
    \textsuperscript{4}Department of Physics, Technical University Dortmund, Otto-Hahn-Straße 4a, 44227 Dortmund, Germany\\
    \small \itshape 
    \textsuperscript{5}German Cancer Consortium (DKTK), Partner Site Essen, Hufelandstraße 55, 45147 Essen, Germany\\
    \small \itshape 
    \textsuperscript{6}Medical Faculty and Faculty of Computer Science, University of Duisburg-Essen, 45141 Essen, Germany\\
    \small \itshape 
    \textsuperscript{7}German Cancer Research Center (DKFZ), 69120 Heidelberg, Germany\\
    \small \itshape 
    \textsuperscript{8}Department of Radiology, University Hospital Essen, Hufelandstraße 55, 45147 Essen, Germany \\
    \small \itshape 
    \textsuperscript{9}Department of Neurosurgery and Spine Surgery, University Hospital Essen, Hufelandstraße 55, 45147 Essen, Germany
}
\begin{document}
\maketitle
\input{sec/0_abstract}    
\input{sec/1_intro}
\input{sec/2_relatedworks}
\input{sec/3_methodology}
\input{sec/4_experiments}
\input{sec/5_results}
\input{sec/6_discussion}
{
    \small
    \bibliographystyle{ieeetr}
    \bibliography{main}
}

\input{sec/X_suppl}

\end{document}

%% file: sec/0_abstract.tex
\begin{abstract}
Deep learning applications in Magnetic Resonance Imaging (MRI) predominantly operate on reconstructed magnitude images, a process that discards phase information and requires computationally expensive transforms. Standard neural network architectures rely on local operations (convolutions or grid-patches) that are ill-suited for the global, non-local nature of raw frequency-domain (k-Space) data. In this work, we propose a novel complex-valued Vision Transformer (kViT) designed to perform classification directly on k-Space data. To bridge the geometric disconnect between current architectures and MRI physics, we introduce a radial k-Space patching strategy that respects the spectral energy distribution of the frequency-domain. Extensive experiments on the fastMRI and in-house datasets demonstrate that our approach achieves classification performance competitive with state-of-the-art image-domain baselines (ResNet, EfficientNet, ViT). Crucially, kViT exhibits superior robustness to high acceleration factors and offers a paradigm shift in computational efficiency, reducing VRAM consumption during training by up to 68$\times$ compared to standard methods. This establishes a pathway for resource-efficient, direct-from-scanner AI analysis.
\end{abstract}

%% file: sec/1_intro.tex
\section{Introduction}
\label{sec:intro}

In today's clinical practice, magnetic resonance imaging (MRI) is a crucial tool for diagnosis and monitoring of various medical conditions. Artificial Intelligence (AI) has shown great potential in enhancing MRI analysis in a wide range of applications, including image segmentation and disease detection~\cite{eggerMedicalDeepLearning2022b, bouhafraDeepLearningApproaches2025}. The current paradigm relies predominantly on processing magnitude-only images in the spatial domain. This approach ignores the fundamental reality of MRI acquisition: data is captured in the frequency domain, known as k-Space. The standard workflow involves reconstructing k-Space data into the image domain via the Inverse Fourier Transform (IFT) before feeding it into neural networks. This conversion process is not merely a formatting step; it is often lossy and computationally expensive. The reconstruction typically discards the phase component of the complex-valued signal~\cite{heckelDeepLearningAccelerated2024a}. Even though the phase information is not easily interpretable by humans, it has been shown to be beneficial for AI-based MRI analysis~\cite{rempeTumorLikelihoodEstimation2024b, liClassificationRegressionSegmentation2024a}. Leveraging the full potential of k-Space data for AI applications requires specialized neural network architectures that can effectively process complex-valued inputs. Recent advancements in complex-valued neural networks have demonstrated promising results in various domains, including MRI~\cite{douMRIDenoisingNonblind2025}, but also in other fields such as Synthetic Aperture Radar (SAR) imaging~\cite{baiTOPSspeedComplexvaluedConvolutional2025} and wireless communication~\cite{yuComplexvaluedNeuralnetworkbasedFederated2024}. 

Another factor to consider when working with MRI data is the common use of undersampling techniques to reduce scan times by only acquiring a subset of the full k-Space data. While there are various algorithms to reconstruct images from undersampled k-Space data~\cite{heckelDeepLearningAccelerated2024a}, these methods can introduce artifacts and loss of information. To mitigate the risk of information loss due to reconstruction, as well as to reduce computational costs, there have been recent efforts to develop AI models that operate directly on undersampled k-Space data~\cite{rempeKstripNovelSegmentation2024c, yenAdaptiveSamplingKspace2024a}. 

However, processing k-Space data presents a unique architectural challenge. Unlike natural images, where pixel dependencies are local, k-Space exhibits a non-local property: a single point in the frequency domain contributes global information to the entire spatial image. Most of today's state-of-the-art (SOTA) neural networks rely on local operations, such as convolutions or patch-based attention mechanisms~\cite{zhaoReviewConvolutionalNeural2024, takahashiComparisonVisionTransformers2024}, to process the input data. We hypothesize that these local operations may not be optimal for capturing the global dependencies present in k-Space data. Vision Transformers (ViTs)~\cite{dosovitskiyImageWorth16x162020} have shown great success in various computer vision tasks due to their ability to model long-range dependencies through patch-based self-attention mechanisms. The idea is to split the input image into smaller squared patches and compute self-attention between these patches. There have been several works proposing different ways of constructing patches for ViTs, such as shifted patches~\cite{liuSwinTransformerHierarchical2021}, deformable patches~\cite{chenDptDeformablePatchbased2021} and even radial patches to mimic the human visual system~\cite{athwaleDarswinDistortionAware2023}. Nevertheless, all these approaches only consider (real-valued) image domain data. 

To bridge the gap between MRI physics and deep learning architecture, we propose a novel Complex-Valued Vision Transformer (kViT) designed specifically for the frequency domain. We rethink the fundamental mechanics of the Vision Transformer, replacing standard grid patches with a radial k-Space Patching strategy. This approach mimics the energy distribution of MRI data, where high-energy contrast information resides in the center and high-frequency edge details occupy the periphery, aligning the network's view with the spectral properties of the signal.

In this work, we present a novel complex-valued transformer architecture specifically designed to process MRI k-Space data. Our main contributions are as follows:
\begin{itemize}
    \item We introduce a fully complex-valued transformer architecture that processes raw, undersampled MRI data directly, effectively utilizing both magnitude and phase information without prior reconstruction.
    \item We propose a physics-informed radial Patching strategy that respects the non-local, spectral structure of k-Space, overcoming the limitations of Cartesian grid patching in the frequency domain.
    \item We demonstrate that our architecture achieves classification performance competitive with state-of-the-art image-domain baselines while reducing VRAM consumption by an order of magnitude (up to $\times$68 reduction).
    \item We provide extensive evaluations on multiple datasets (fastMRI Prostate, Knee, and in-house Glioma), proving the model's robustness to severe undersampling factors where traditional approaches degrade.
\end{itemize}

\begin{figure}[t]
    \centering
    \includegraphics[width=0.9\linewidth]{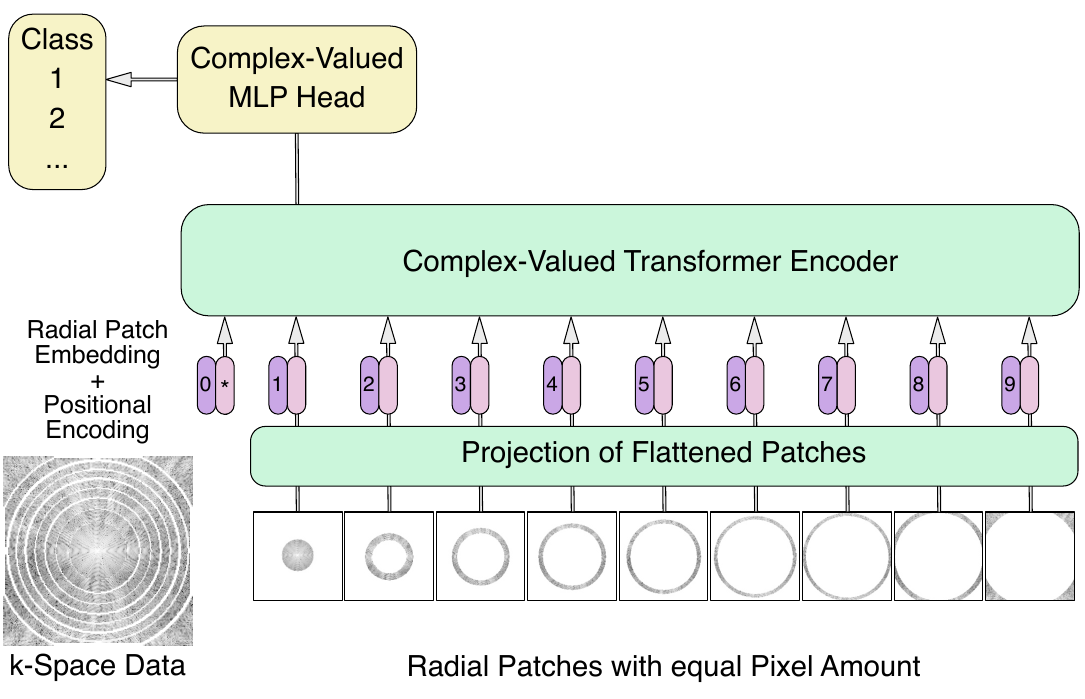}
    \caption{Illustration of the proposed radial k-Space patches and their application within the complex-valued transformer architecture.}
    \label{fig:methodology}
\end{figure}

%% file: sec/2_relatedworks.tex
\section{Related Works}
\label{sec:related_works}

\subsection{(Complex-Valued) Vision Transformers}

One powerful architecture that has gained significant attention in recent years is the ViT. ViTs utilize a self-attention mechanism that allows the model to capture long-range dependencies in the input data. The input image is divided into smaller patches, which are then flattened and linearly embedded before being fed into the transformer layers. This patch-based approach enables the model to learn relationships between different regions of the image, making it particularly effective for tasks such as image classification and segmentation. ViTs also found their way into the domain of medical imaging. Manzari et al. propose MedViT, a transformer-based architecture for medical image classification that outperforms traditional convolutional neural networks (CNNs) on several benchmark datasets~\cite{manzariMedViTRobustVision2023}. Joy et al. utilize a modified ViT architecture for Alzheimer Detection in MRI images~\cite{joyViTADLeveragingModified2025}. There have also been approaches to translate the ViT architecture in the complex domain. Alkhatib uses a complex-valued ViT for radar target recognition in~\cite{alkhatibPolSARImageClassification2025}, while Li et al. propose a complex-valued ViT for automatic modulation recognition~\cite{liComplexvaluedTransformerAutomatic2024}. These works demonstrate the potential of complex-valued ViTs in various applications, but to the best of our knowledge, there has been no prior work exploring complex-valued ViTs for MRI k-Space data.

\subsection{MRI Raw Data}

MRI raw data, also known as k-Space data, is the frequency domain representation of the spatial image and the original form in which MRI data is acquired. Each point in k-Space contains information about the entire image and is complex-valued, consisting of both magnitude and phase components. While the k-Space magnitude can take on arbitrary non-negative values, the phase component is typically represented as values within the range of $[-\pi, \pi]$. The Inverse Fourier Transform (IFT) is used to convert the k-Space data into the spatial domain. This process is reversible, however most downstream tasks discard the phase component, only using the magnitude information. One unique property of this space is that distances and relationships between points in k-Space do not directly correspond to those in the spatial domain. For example, two points that are close together in k-Space may represent distant regions in the spatial image, and vice versa. This nature of k-Space data presents challenges for traditional machine learning algorithms, which often assume local image-information. K-Space points in the center of the frequency domain contain low-frequency information that captures the overall structure and contrast of the image, while points in the outer regions contain high-frequency information that captures fine details and edges. 

The research field of utilizing this raw information for downstream tasks is still in its infancy. While the use of k-Space data for reconstruction tasks has been extensively studied, its application in other areas such as image classification and segmentation is relatively unexplored. Yen et al. propose the ASMR architecture which utilizes reinforced learning to adaptively sample the k-Space to optimize further classification tasks~\cite{yenAdaptiveSamplingKspace2024a}. Multiple works were already able to show that utilizing k-Space data for classification at high undersampling rates can outperform image-domain methods~\cite{rempeTumorLikelihoodEstimation2024c, liClassificationRegressionSegmentation2024a}. Undersampling is a common technique used in MRI to reduce acquisition time by acquiring fewer k-Space samples. However, this can lead to artifacts and loss of information in the reconstructed images. By directly utilizing the k-Space data, these methods can potentially mitigate the negative effects of undersampling and improve classification performance. Despite these promising results, there is still a lack of comprehensive studies exploring the full potential of k-Space data for various downstream tasks. Further research is needed to fully understand the advantages and limitations of using k-Space data in different applications.

%% file: sec/3_methodology.tex
\section{Methodology}

In this work we propose a novel complex-valued Vision Transformer (ViT) architecture specifically designed to operate directly on (MRI) k-Space data, utilizing both magnitude and phase information, as well as its non-Euclidean properties. In a first step, we implement each component of the ViT architecture in the complex domain, including complex-valued multi-head self-attention and complex-valued feed-forward networks. To receive a real-valued output from the complex-valued ViT, the final output of the complex-valued linear layer is split into its real and imaginary parts, which are then averaged. 

Due to its non-local nature, using a traditional patching strategy is less intuitive and as our ablation studies show, impacting performance negatively. Therefore, we introduce a novel radial patching strategy, which splits the k-Space data into radial patches, preserving the inherent structure of the data. An overview of our proposed methodology is illustrated in Fig.~\ref{fig:methodology}. In the following we will explain the radial patching approach in more detail.

\subsection{Radial k-Space Patching}

Traditional ViTs rely on splitting the input image into smaller squared patches, which are then flattened and linearly projected into a lower-dimensional embedding space. Instead of using squared patches, we propose to split the k-Space data into radial patches, going from the center of k-Space outwards. To then be able to use these radial patches within the ViT architecture, we have to ensure that each patch has the same number of pixels. Therefore, we define a fixed number of pixels per radial patch, denoted as $P$. The core idea is to rank all pixels in the k-Space grid based on their radial distance $r(x,y)=\sqrt{x^2+y^2}$ from the center, resulting in a list of radii $R$. Once ranked, this 1D list of pixels is divided into $N$ contiguous segments of equal size. The equal distribution of pixels per patch is ensured by choosing $N$ such that $N=\frac{H\cdot W}{P}$, where $H$ and $W$ are the height and width of the k-Space grid, respectively. Each of these segments then forms one radial patch. Optionally, the patches are then weighted with learnable complex-valued weights to further enhance the representation capability of each patch. Finally, each radial patch is flattened and projected into the embedding space using a complex-valued linear layer, similar to traditional ViTs.

\subsection{Complex-Valued Positional Embedding}

To retain positional information of each radial patch within the ViT architecture, we introduce a complex-valued positional embedding. Unlike traditional ViTs that use learnable real-valued positional embeddings, we propose two different strategies for complex-valued positional embeddings: (1) a learnable complex-valued positional embedding, where a single complex-valued vector is learned for each radial patch during training, and (2) an adaptation of the Rotary Position Embedding (RoPE)~\cite{suRoformerEnhancedTransformer2024, heoRotaryPositionEmbedding2024}. RoPE introduces relative positional information by rotating the query and key vectors in the complex plane based on their positions. This is particularly suitable for our complex-valued ViT, as it naturally operates in the complex domain. In our experiments, we evaluate both strategies to determine their effectiveness in capturing positional information in k-Space data.

\subsection{Complex-Valued ViT Architecture}

The overall architecture of our complex-valued ViT is similar to traditional ViTs, consisting of multiple layers of complex-valued multi-head self-attention and complex-valued feed-forward networks. Each layer is followed by layer normalization and residual connections, adapted for the complex domain. The output of the final transformer layer is then passed through a complex-valued linear layer to produce the final output, which is then split into its real and imaginary parts and averaged to obtain a real-valued prediction.

%% file: sec/4_experiments.tex
\section{Experiments}

To evaluate the performance of our proposed complex-valued Vision Transformer (ViT) architecture with radial k-Space patching and complex-valued positional embeddings, we conduct a series of experiments on publicly available MRI dataset, as well as an in-house dataset. For each dataset, we also evaluate the performance of our method under different levels of undersampling in k-Space, simulating various acceleration factors commonly used in MRI acquisition. We compare our method against several baselines, including traditional real-valued ViTs and convolutional neural networks (CNNs). 

\subsection{Datasets and Preprocessing}

In total we use three different datasets for our experiments: the fastMRI prostate dataset~\cite{tibrewalaFastMRIProstatePublic2024b}, the fastMRI knee dataset~\cite{knollFastMRIPubliclyAvailable2020a} in combination with the fastMRI+ labels~\cite{zhaoFastMRIClinicalPathology2022}, and an in-house brain MRI dataset.

From the fastMRI prostate dataset, we use axial T2-weighted scans from 226 patients, including 41 patients for a hold-out test set, with an in-plane resolution of $0.56\times0.56$ mm at a field strength of 3 T. In total we use 6890 2D slices for training and validation, and 1245 2D slices for testing. The data is labeled based on their PI-RADS score~\cite{turkbeyProstateImagingReporting2019b}, which is a clinical standard for assessing the risk of prostate cancer. Following the implementation of the authors, we combine PI-RADS scores 1-2 into a low-risk class and PI-RADS scores 3-5 into a high-risk class, resulting in a binary classification task.

For the fastMRI knee dataset, we use the single-coil coronal PD-weighted scans from 1398 patients, scanned at 1.5 and 3 T with a slice thickness of 3 mm. In total we use 46698 2D slices for training and validation, and 1752 2D slices for testing. Multiple pathologies are labeled in the fastMRI+ dataset. We focus on the detection of meniscal tears and ACL sprains. Additionally we classify for a general abnormality label, indicating the presence of any pathology in the knee joint, leading to 4-class classification task. 

The in-house brain MRI dataset consists of 432 patients scanned at 1.5 T and 3 T with a variety of transverse MRI sequences, including T1-weighted, T2-weighted, and FLAIR scans with different in-plane resolutions and slice thicknesses [24-11872-BO]. This dataset only contains patient-level labels, including Glioma, Metastasis, and Healthy controls. Due to the patient-level labels, we perform a multiple instance learning (MIL) approach \cite{dietterichSolvingMultipleInstance1997}, where we aggregate the slice-level predictions to obtain patient-level predictions. Another reason for this approach is that for one patient, multiple scans with different sequences might be available. In total we use 14803 2D slices. 360 patients are used for training and validation, and 72 patients for testing.

For all experiments we perform 5-fold cross-validation and test each fold on the hold-out test set. During training, we randomly split the training data into 80\% training and 20\% validation data, while ensuring that no patient appears in both sets.

Before training, we preprocess the data by standardizing the data slice-wise to zero mean and unit variance. For augmentation, we apply standard image augmentations, including horizontal flips, random rotations and skewing. In case of the complex-valued k-Space data, we first transform the data to image space, apply the augmentations to real and imaginary parts separately, and then transform the data back to k-Space. We also apply random cutout directly on the k-Space data, by setting random rectangular regions in k-Space to zero. To apply the same augmentations to all baselines, we transform the image-space data into the k-Space and apply the cutout before transforming the data back to image space for the real-valued baselines.

\subsection{Evaluation Metrics}

We report the area under the receiver operating characteristic curve (AUROC) as well as the area under precision-recall curve (AUPRC) as our main evaluation metrics for all classification tasks. We report the average and standard deviation of these metrics across the 5 cross-validation folds. We also report these metrics for different undersampling factors $R$ in k-Space, simulating acceleration factors of 2x, 4x, 6x, 8x, 12x, 16x, and 24x. For each acceleration factor, we keep different center fractions of fully sampled k-Space, ranging from 8\% for $R=4$ to 0.8\% for $R=24$. A detailed overview of the used acceleration factors and center fractions is provided in Table~\ref{tab:undersampling_rates}.

We compare our model with the following baselines:
\begin{itemize}
    \item \textbf{Real-valued ViT}: A standard real-valued Vision Transformer architecture~\cite{dosovitskiyImageWorth16x162020}, similar to our proposed method, but operating on real-valued image-space data. We use the ViT-Tiny architecture 
    \item \textbf{EfficientNet}: A convolutional neural network (CNN) architecture, which has shown strong performance on various image classification tasks. We use the EfficientNet-B0 architecture~\cite{tanEfficientnetRethinkingModel2019}, which is a lightweight version of EfficientNet.
    \item \textbf{ResNet}: A standard ResNet architecture~\cite{heDeepResidualLearning2016a}. We use the ResNet-50 architecture.
\end{itemize}

\subsection{Implementation Details}

All baselines are implemented via the \textit{Timm} library~\cite{rw2019timm}. For better comparability, we do not use pretrained weights for any of the models and train all models from scratch. All baselines showed optimal performance with a learning rate of $1\cdot10^{-4}$ using the AdamW optimizer \cite{loshchilovDecoupledWeightDecay2017a} and a batch size of 128. For the proposed complex-valued ViT we used a learning rate of $1\cdot10^{-4}$ with the AdamW optimizer and a batch size of 64. For the MIL approach, we use a batch size of 1 due to memory constraints. We perform early stopping based on the validation loss with a patience of 15 epochs. For all experiments, the Cross Entropy loss is used as the loss function, with a weighted factor to account for class imbalance in case of the Prostate dataset.

Our proposed method is trained with the following hyperparameters: For the fastMRI prostate and knee dataset we use 6 Transformer layers with 16 attention heads and an embedding dimension of 256. The MLP dimension is set to 768. We use a dropout rate of 0.1 in all Transformer layers. The input k-Space is split into 16 rings with a maximum of 3800 pixels each.
For the in-house dataset in the MIL setting we reduce the transformer size to use 2 Transformer layers with 4 attention heads and an embedding dimension of 48. The MLP dimension is set to 92.

All experiments are conducted on a single NVIDIA A100 GPU with 80GB of memory. We make our code publicly available at: \url{https://github.com/TIO-IKIM/kViT}.

%% file: sec/5_results.tex
\section{Results}
\label{sec:results}

In this section, we present the results of our proposed method on various benchmark datasets. We compare our approach with state-of-the-art (SOTA) methods and demonstrate its effectiveness throughout different undersampling rates.

\subsection{FastMRI Prostate}

\begin{figure}[htbp]
    \centering
    \includegraphics[width=1\linewidth]{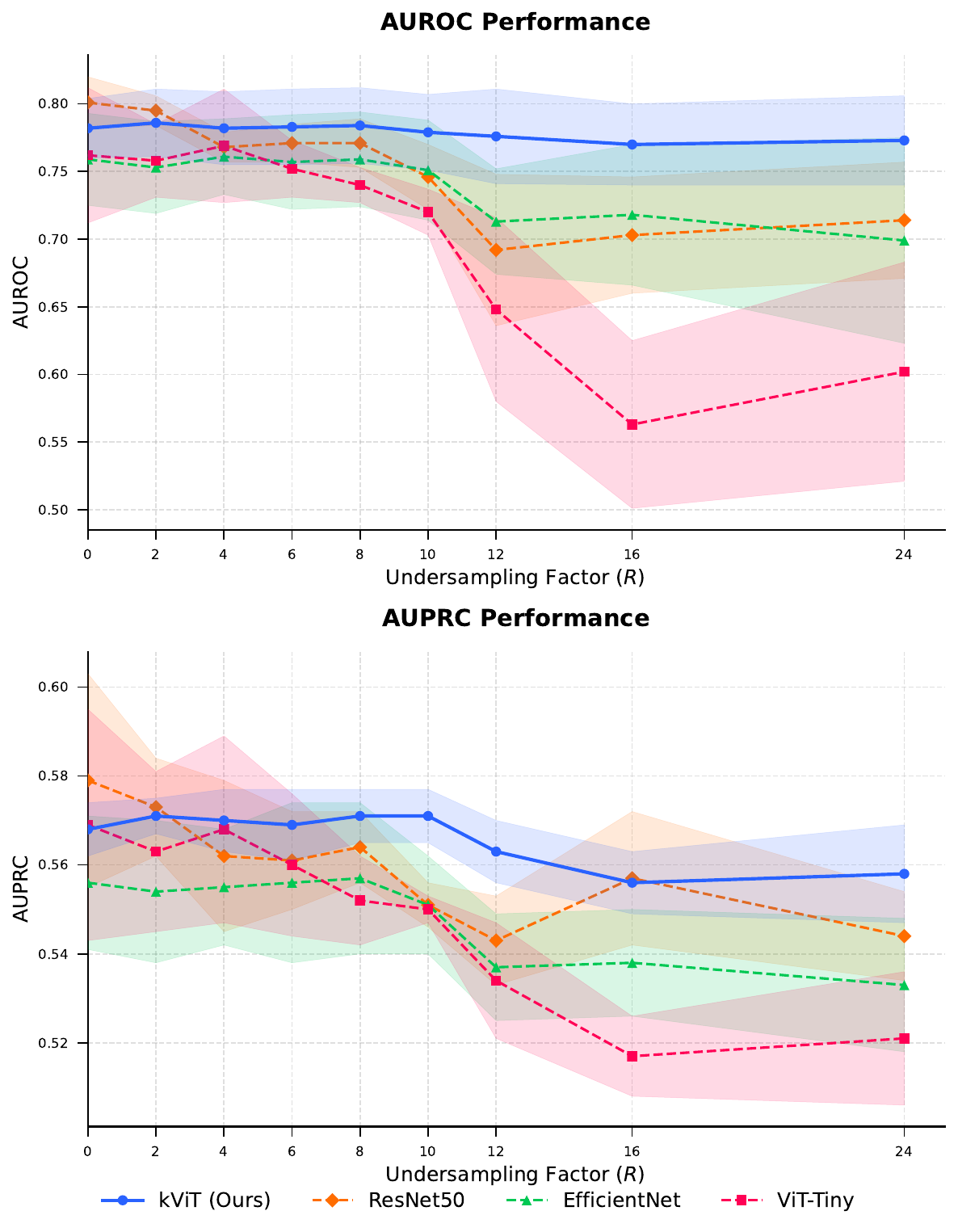}
    \caption{Classification results on the fastMRI prostate dataset. Note the consistent performance at higher acceleration factors. The shown results are the mean and standard deviation over five folds tested on a hold-out test set.}
    \label{fig:prostate_results}
\end{figure}

\begin{figure}[htbp]
    \centering
    \includegraphics[width=1\linewidth]{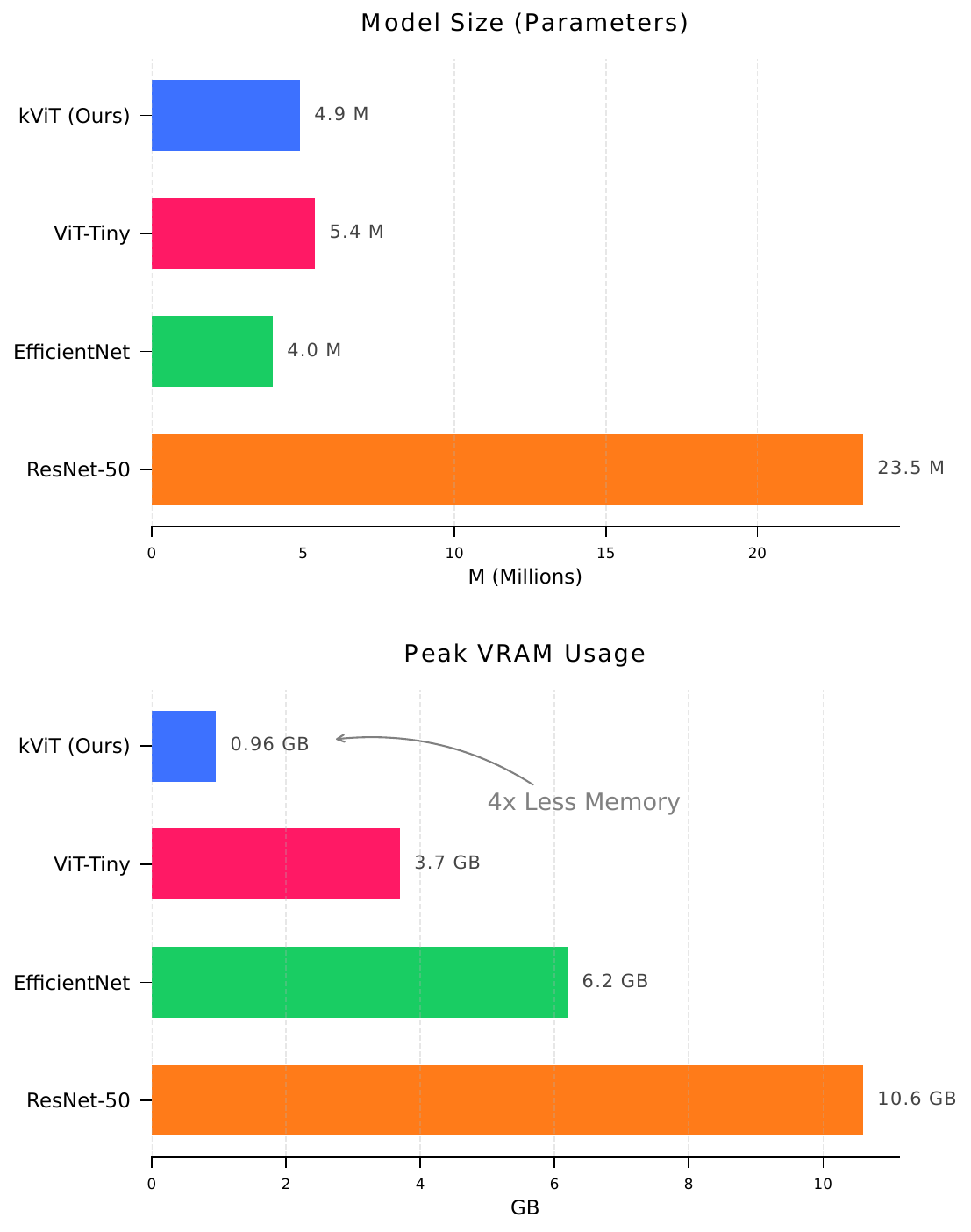}
    \caption{The number of parameters and VRAM consumption during training for different model configurations on the fastMRI prostate and knee dataset at a batch size of 64. The proposed method achieves competitive performance with significantly reduced VRAM usage.}
    \label{fig:vram_prostate}
\end{figure}

Fig.~\ref{fig:prostate_results} illustrates the classification accuracy of our method on the fastMRI prostate dataset across various undersampling rates. Our approach consistently outperforms existing methods at higher acceleration factors, demonstrating its robustness and effectiveness in challenging scenarios. While the proposed method achieves comparable results to the baselines at lower acceleration factors, it excels as the undersampling rate increases, highlighting its capability to recover critical information even with limited data. While the kViT achieves an AUROC of $78.2\pm2.2$ on fully sampled data, our method attains an AUROC of $77.0\pm3.0$ at an acceleration factor of $16\times$, while the best baseline at low undersampling rates, ResNet50, drops from $80.1\pm1.9$ to $70.3\pm4.3$ under the same conditions. At the same time, the proposed method requires significantly less VRAM during training, as shown in Fig.~\ref{fig:vram_prostate}. While kViT requires only 0.96GB of VRAM with 4.9 million parameters during training, ResNet50 used 10.6GB of VRAM with 23.5 million parameters. The baseline with the least VRAM consumption, ViT-Tiny, requires 3.7GB of VRAM with 5.4 million parameters, but drops from $76.2\pm5.0$ AUROC on fully sampled data to $56.3\pm6.2$ at an acceleration factor of $16\times$. All results can be found in Tab.~\ref{tab:prostate_results}.

On this dataset, the RoPE position embedding in combination with the additional scaling parameters proved to be more effective than the learnable position embedding, likely due to the higher generalization capabilities of the RoPE embedding. Further ablation studies will be presented in Sec.~\ref{sec:ablation_studies}.

\subsection{FastMRI Knee}

\begin{figure}[htbp]
    \centering
    \includegraphics[width=1\linewidth]{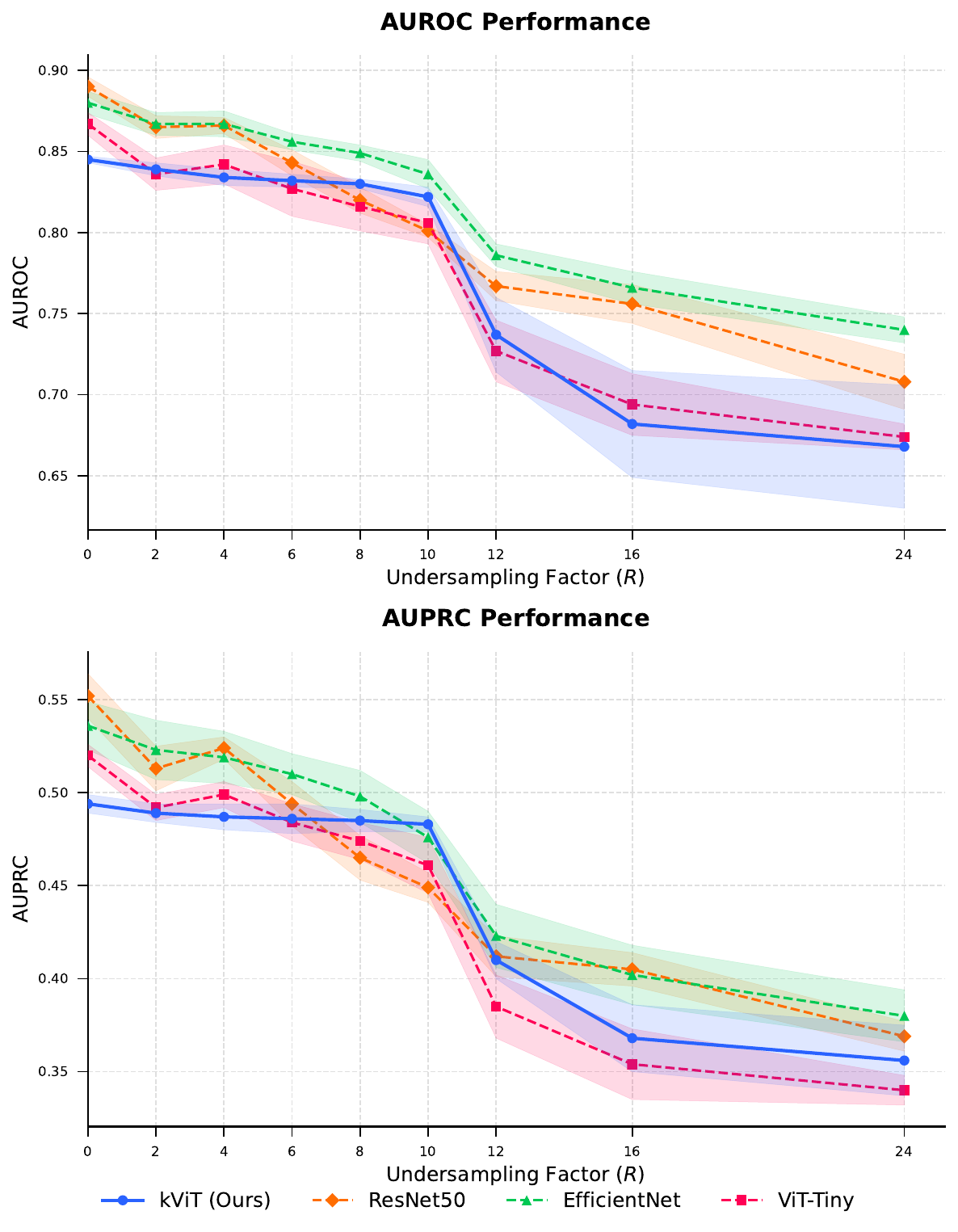}
    \caption{Classification results on the fastMRI knee dataset. The proposed model reaches comparable performance to the SOTA at reduced VRAM consumption, but showing the same decline in performance at higher acceleration factors as the baselines. The shown results are the mean and standard deviation over five folds tested on a hold-out test set.}
    \label{fig:knee_results}
\end{figure}

Fig. \ref{fig:knee_results} shows the classification results on the fastMRI knee dataset. Here, our method achieves comparable performance to the SOTA methods at reduced VRAM consumption. However, similar to the baselines, our approach exhibits a decline in performance at higher acceleration factors. While kViT achieves lower performance on fully sampled data with an AUROC of $84.5\pm0.2$ compared to ResNet50's $89.0\pm0.6$, it maintains better performance at an acceleration factor of $10\times$ with an AUROC of $82.2\pm0.6$ compared to ResNet50's $80.1\pm0.4$. At high accelerations the proposed method's performance drops similar to all baselines. In Sec.~\ref{sec:discussion} we will discuss potential reasons for this behavior in comparison to the prostate dataset. All results can be found in Tab.~\ref{tab:knee_results}.

\begin{figure}[htbp]
    \centering
    \includegraphics[width=1\linewidth]{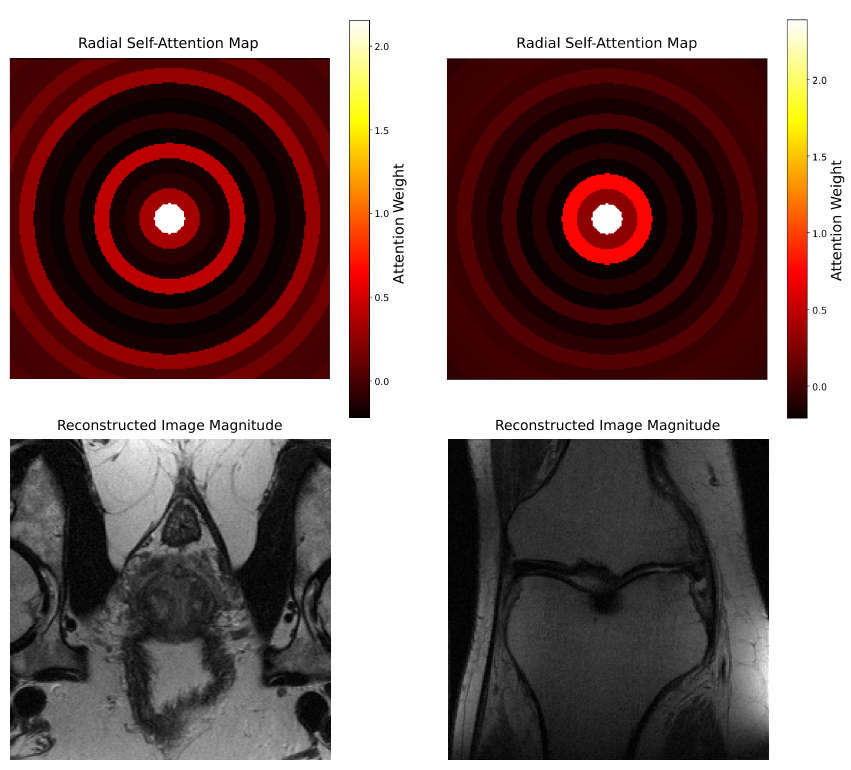}
    \caption{Attention Maps from the last Transformer layer of the proposed method on the (left) fastMRI prostate dataset and (right) fastMRI knee dataset and the corresponding input data transformed into the image space magnitude. The attention maps are visualized logarithmically for better visibility. The model focuses primarily on the center region of k-Space, while also attending to high-frequency components in the outer regions.}
    \label{fig:attn_maps}
\end{figure}

Fig. \ref{fig:attn_maps} visualizes attention maps from the last Transformer layer of the proposed method on both the fastMRI prostate and knee datasets. The attention maps reveal that the model predominantly focuses on the center region of k-Space, which contains low-frequency information. However, it also attends to high-frequency components in the outer regions of k-Space, indicating that the model effectively captures important details necessary for accurate classification.

\subsection{Multiple Instance Learning}

This experiment evaluates the proposed method in a multiple instance learning (MIL) setting on the fastMRI prostate dataset. One challenge is the small number of patients in the dataset, which can quickly lead to overfitting. For this reason we replaced the RoPE position embedding in combination with the scaling parameters with the simpler learnable position embedding. Additionally we replaced the one-layer classification head with a multi-layer perceptron (MLP) with one hidden layer of size 128 and ReLU activation as well as dropout with a rate of 0.2 to allow the model to better process the aggregated features from multiple slices.

\begin{figure}[htbp]
    \centering
    \includegraphics[width=1\linewidth]{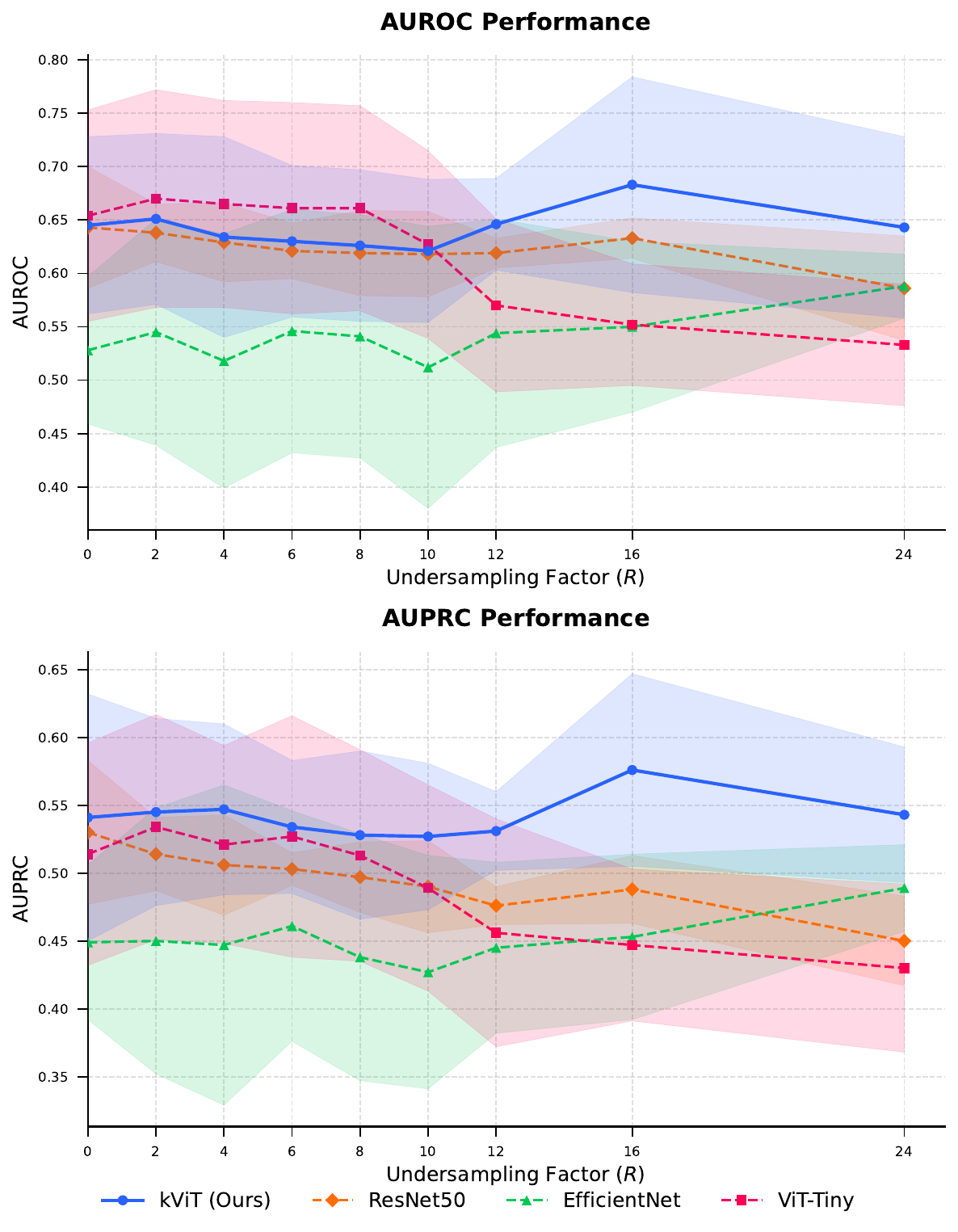}
    \caption{Results on the in-house dataset in a multiple instance learning setting. While the AUROC is similar for all methods, the proposed method shows higher AUPRC over all undersampling rates. The shown results are the mean and standard deviation over five folds tested on a hold-out test set.}
    \label{fig:mil_results}
\end{figure}

The results in Fig.~\ref{fig:mil_results} show that while the AUROC is similar for all methods, the proposed method shows higher AUPRC over all undersampling rates. kViT achieves a consistently higher AUPRC for all undersampling rates, with an AUPRC of $54.1\pm9.1$ on fully sampled data and $54.3\pm5.0$ at an acceleration factor of $24\times$. In comparison, the real-valued ViT reaches an AUPRC of $51.4\pm8.2$ on fully sampled data and drops to $43.0\pm6.2$ at an acceleration factor of $24\times$. All results can be found in Tab.~\ref{tab:mil_results}.

\begin{figure}[htbp]
    \centering
    \includegraphics[width=1\linewidth]{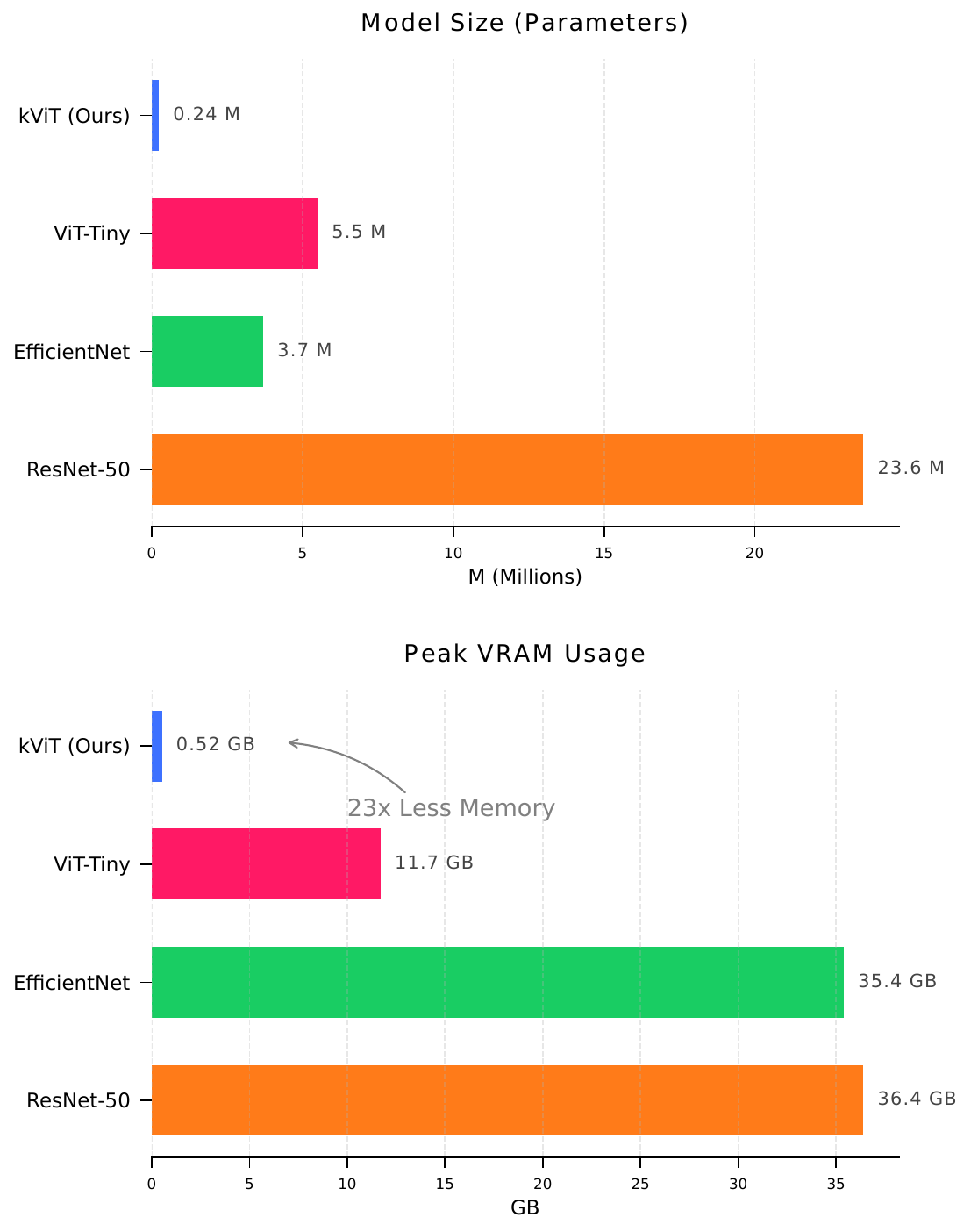}
    \caption{The number of parameters and VRAM consumption during training for different model configurations on the in-house dataset at a MIL batch size of 1. The proposed method achieves better performance with significantly reduced VRAM usage.}
    \label{fig:vram_mil}
\end{figure}

However more significant is the difference in VRAM consumption during training, as shown in Fig.~\ref{fig:vram_mil}. While the proposed method only requires 0.52GB of VRAM with 24.3 million parameters at a MIL batch size of 1, the next smallest baseline used in our experiments, ViT-Tiny, requires 11.7GB of VRAM with 5.5 million parameters. This results in a reduction of VRAM consumption by a factor of 23 when using the proposed method. The next larger baseline, EfficientNet-B0, requires 35.4GB of VRAM with 3.7 million parameters. Using the proposed method thus decreases the VRAM consumption by a factor of 68. This significant reduction in VRAM usage allows for training with larger MIL batch sizes or higher resolution images, while using smaller hardware resources.

Fig.~\ref{fig:mil_slice_importance} and Fig.~\ref{fig:mil_attn_map} visualize the slice importance scores and attention maps from the last Transformer layer of the proposed method in the MIL setting. The slice importance scores indicate which slices the model considers most relevant for the final patient-level classification. The model weights the existing sequences differently, focusing more on the $T_1$ sequence. The attention map looks different compared to the attention maps of the models trained on the fastMRI datasets. Here, the model attends more to the high-frequency components in the outer regions of k-Space, which may be due to the nature of the pathology being detected in this dataset.

\subsection{Ablation Studies}
\label{sec:ablation_studies}

We conduct a series of ablation studies to evaluate the impact of different design choices on the performance of our proposed method. Due to training time constraints, these studies were performed on the fastMRI prostate dataset. We again report all results as an average of five folds tested on a hold-out test set. The results of these studies are summarized in Tab.~\ref{tab:ablation_undersampling}.

\begin{table*}[htbp]
    \centering
    \caption{Ablation study varying the undersampling ratio (None, 8x, 16x). We report AUROC and AUPRC for each scenario.}
    \label{tab:ablation_undersampling}
    \setlength{\tabcolsep}{5pt} 
    \begin{tabular}{l ccc c ccc}
        \toprule
        & \multicolumn{3}{c}{\textbf{AUROC ($\pm$ std)}} & & \multicolumn{3}{c}{\textbf{AUPRC ($\pm$ std)}} \\
        \cmidrule(lr){2-4} \cmidrule(lr){6-8}
        \textbf{Model Variant} & \textbf{No US} & \textbf{8x} & \textbf{16x} & & \textbf{No US} & \textbf{8x} & \textbf{16x} \\
        \midrule
        \textbf{Full Method} & \textbf{0.782 {\textpm} 0.022} & \textbf{0.784 {\textpm} 0.028} & \textbf{0.770 {\textpm} 0.030} & & \textbf{0.568 {\textpm} 0.006} & \textbf{0.571 {\textpm} 0.006} & \textbf{0.556 {\textpm} 0.007} \\
        \midrule
        \quad 8 rings       & 0.750 {\textpm} 0.017 & 0.757 {\textpm} 0.014 & 0.718 {\textpm} 0.021 & & 0.560 {\textpm} 0.006 & 0.565 {\textpm} 0.009 & 0.546 {\textpm} 0.007 \\
        \quad 32 rings      & 0.762 {\textpm} 0.028 & 0.764 {\textpm} 0.025 & 0.731 {\textpm} 0.029 && 0.558 {\textpm} 0.007 & 0.559 {\textpm} 0.007 & 0.541 {\textpm} 0.007 \\
        \quad w/o RoPE      & 0.773 {\textpm} 0.025 & 0.784 {\textpm} 0.024 & 0.769 {\textpm} 0.019 && 0.563 {\textpm} 0.008 & 0.570 {\textpm} 0.011 & 0.554 {\textpm} 0.008 \\
        \quad w/o Cutout Aug& 0.750 {\textpm} 0.023 & 0.755 {\textpm} 0.031 & 0.732 {\textpm} 0.029 && 0.555 {\textpm} 0.006 & 0.558 {\textpm} 0.009 & 0.544 {\textpm} 0.007 \\
        \quad w/o Phase     & 0.685 {\textpm} 0.024 & 0.670 {\textpm} 0.030 & 0.551 {\textpm} 0.070 && 0.536 {\textpm} 0.007 & 0.531 {\textpm} 0.008 & 0.512 {\textpm} 0.010 \\
        \bottomrule
    \end{tabular}
\end{table*}

The ablation studies indicate that the number of k-Space rings used for patching has a significant impact on the model's performance. In all our experiments, 16 rings provided the best results. Another important insight is the necessity of incorporating phase information in the input data. Removing the phase information led to a substantial drop in performance, especially at higher undersampling rates. This highlights the importance of utilizing the full complex-valued nature of MRI data for optimal classification performance. Additionally, the use of the Cutout Augmentation technique contributed positively to the model's robustness, as its removal resulted in decreased performance across all undersampling rates. Lastly, while the RoPE positional embedding improved performance slightly, its impact was less pronounced compared to other factors.

%% file: sec/6_discussion.tex
\section{Discussion \& Conclusion}
\label{sec:discussion}

In this work, we proposed kViT, a novel complex-valued vision transformer architecture for MRI classification tasks directly from (undersampled) k-Space data. The proposed method leverages radial k-Space patching and complex-valued positional embeddings to effectively capture the inherent structure of k-Space data. This approach proved to be particularly advantageous in scenarios with high undersampling factors, where traditional real-valued models often struggle. Our experiments on multiple MRI datasets demonstrated that kViT consistently outperforms or matches real-valued baselines while requiring significantly less VRAM during training and inference. This efficiency gain is particularly important in medical imaging applications, where high-resolution data and complex models can lead to substantial computational demands.

One interesting observation from our experiments is the differing performance trends between the prostate and knee MRI datasets. While kViT excelled in the prostate dataset, especially at high undersampling rates, its performance on the knee dataset was comparable to real-valued baselines but did not exhibit the same robustness to undersampling. We hypothesize that this discrepancy may be attributed to the nature of the pathologies present in the datasets. On T2-weighted MRI, prostate cancer often appears as a focal region of hypointensity with a different texture than the surrounding healthy gland. This change in cellular density leads to a distinct alteration in the energy spectrum~\cite{wibmerHaralickTextureAnalysis2015}, and the phase in this case can give additional information due to susceptibility changes.
In contrast, the pathologies in the knee MRI dataset, such as meniscal tears and ACL sprains, are geometric discontinuities~\cite{desmetUseTwoslicetouchRule2006}. Image domain models, especially CNNs, are particularly well-suited to capture such geometric features due to their localized receptive fields and ability to learn spatial hierarchies. Consequently, the advantages of complex-valued representations in k-Space may be less pronounced for these types of pathologies.
To prove this hypothesis, further investigations are needed, potentially involving additional datasets and pathology types. If true, the proposed approach should excel in tasks where the pathology induces changes in the energy spectrum rather than geometric changes, such as fat and iron quantification in the liver, or differentiation of hemorrhagic and non-hemorrhagic stroke lesions. 

The proposed kViT architecture opens several possibilities for further research. In this work we did not explore hybrid approaches that combine k-Space and image-space information. Future work could investigate architectures that process both domains, potentially leveraging the strengths of each. Additionally, we did not pretrain our models on large-scale datasets, which has been shown to be beneficial for real-valued ViTs. While data scarcity is a challenge in the domain of MRI raw data, there have been recent efforts to circumvent this issue~\cite{rempePhaseGenDiffusionBasedApproach2025a}.

Another limitation of our current approach is the "ring-only" approach presented in the radial k-Space patching strategy. While this method proved efficient and effective, it may be one reason for the worse performance on the knee dataset. Because we patch the k-Space into full rings, spatial information about specific directions in k-Space is lost. Future work sees the advanced implementation of a polar patching strategy, including rings and sectors, which could help retain more spatial information in k-Space. Additionaly, weighting the patches based on their distance to the center of k-Space could further improve the model's ability to focus on important regions in k-Space.

In conclusion, kViT represents a significant step forward in the application of complex-valued deep learning models for MRI classification tasks. By directly operating on k-Space data and effectively handling undersampling, kViT has the potential to enhance diagnostic accuracy while reducing computational requirements. Future research should continue to explore the capabilities of complex-valued models in medical imaging, particularly in relation to different pathology types and hybrid domain approaches. One possible use case for kViT could be in pre-screening scenarios, where fast and efficient analysis of undersampled k-Space data could help identify cases that require further detailed imaging, thereby optimizing MRI scanner usage and reducing patient wait times.

\section*{Impact Statement}

This work advances efficient medical imaging by enabling classification directly from raw, undersampled k-Space data. By reducing VRAM usage by an order of magnitude and maintaining performance at high acceleration factors, our method promotes "Green AI" and enables the deployment of diagnostic models on resource-limited hardware. Additionally, robustness to undersampling supports faster MRI acquisition protocols, potentially improving patient comfort and optimizing hospital workflows.

\section*{Acknowledgement}

This work received funding from the the Bruno \& Helene Jöster Foundation, KITE (Plattform für KI-Translation Essen) from the REACT-EU initiative (https://kite.ikim.nrw/) and the Cancer Research Center Cologne Essen (CCCE).
The authors acknowledge that this manuscript was edited with the assistance of LLMs. The authors declare no competing interests.

%% file: sec/X_suppl.tex
\clearpage
\setcounter{page}{1}
\maketitlesupplementary
\renewcommand{\thetable}{A\arabic{table}}
\renewcommand{\thefigure}{A\arabic{figure}}

\section{Supplementary Material}
\label{sec:supplementary}

\begin{table*}[htbp]
\centering
\setlength{\tabcolsep}{12pt} 
\renewcommand{\arraystretch}{1.3} 
\caption{Undersampling rates and corresponding center fractions used in all experiments.}
\label{tab:undersampling_rates}
    \begin{tabular}{lc}
    \toprule
    Undersampling Rate & Center Fraction \\
    \midrule
    0   & 1.0 \\
    2   & 0.04 \\
    4   & 0.08 \\
    6   & 0.05 \\
    8   & 0.04 \\
    10  & 0.03 \\
    12  & 0.02 \\
    16  & 0.015 \\
    24  & 0.008 \\
    \bottomrule
    \end{tabular}
\end{table*}

\begin{table*}[htbp]
\centering
\setlength{\tabcolsep}{12pt} 
\renewcommand{\arraystretch}{1.3} 
\caption{Classification results on the fastMRI prostate dataset at different undersampling rates. The proposed method (kViT) consistently outperforms the baselines, especially at higher acceleration factors. Best results are highlighted in \textbf{bold}.}
\label{tab:prostate_results}
    \begin{tabular}{lcccc}
    \toprule
    Undersampling & kViT (Ours) & ResNet50 & EfficientNet & ViT-Tiny \\
    \midrule
    \multicolumn{5}{l}{AUROC (\textpm{} std)} \\
    0  & 0.782 {\textpm} 0.022 & \textbf{0.801 {\textpm} 0.019} & 0.759 {\textpm} 0.034 & 0.762 {\textpm} 0.050 \\
    2  & 0.786 {\textpm} 0.025 & \textbf{0.795 {\textpm} 0.011} & 0.753 {\textpm} 0.034 & 0.758 {\textpm} 0.027 \\
    4  & \textbf{0.782 {\textpm} 0.027} & 0.768 {\textpm} 0.011 & 0.761 {\textpm} 0.028 & 0.769 {\textpm} 0.042 \\
    6  & \textbf{0.783 {\textpm} 0.028} & 0.771 {\textpm} 0.014 & 0.757 {\textpm} 0.035 & 0.752 {\textpm} 0.021 \\
    8  & \textbf{0.784 {\textpm} 0.028} & 0.771 {\textpm} 0.018 & 0.759 {\textpm} 0.035 & 0.740 {\textpm} 0.013 \\
    16 & \textbf{0.770 {\textpm} 0.030} & 0.703 {\textpm} 0.043 & 0.718 {\textpm} 0.052 & 0.563 {\textpm} 0.062 \\
    24 & \textbf{0.773 {\textpm} 0.033} & 0.714 {\textpm} 0.043 & 0.699 {\textpm} 0.076 & 0.602 {\textpm} 0.081 \\
    \midrule
    \multicolumn{5}{l}{AUPRC (\textpm{} std)} \\
    0  & 0.568 {\textpm} 0.006 & \textbf{0.579 {\textpm} 0.024} & 0.556 {\textpm} 0.015 & 0.569 {\textpm} 0.026 \\
    2  & 0.571 {\textpm} 0.004 & \textbf{0.573 {\textpm} 0.011} & 0.554 {\textpm} 0.016 & 0.563 {\textpm} 0.018 \\
    4  & \textbf{0.570 {\textpm} 0.007} & 0.562 {\textpm} 0.017 & 0.555 {\textpm} 0.013 & 0.568 {\textpm} 0.021 \\
    6  & \textbf{0.569 {\textpm} 0.008} & 0.561 {\textpm} 0.011 & 0.556 {\textpm} 0.018 & 0.560 {\textpm} 0.016 \\
    8  & \textbf{0.571 {\textpm} 0.006} & 0.564 {\textpm} 0.008 & 0.557 {\textpm} 0.017 & 0.552 {\textpm} 0.010 \\
    10 & \textbf{0.571 {\textpm} 0.006} & 0.551 {\textpm} 0.005 & 0.551 {\textpm} 0.011 & 0.550 {\textpm} 0.003 \\
    12 & \textbf{0.563 {\textpm} 0.007} & 0.543 {\textpm} 0.010 & 0.537 {\textpm} 0.012 & 0.534 {\textpm} 0.013 \\
    16 & 0.556 {\textpm} 0.007 & \textbf{0.557 {\textpm} 0.015} & 0.538 {\textpm} 0.012 & 0.517 {\textpm} 0.009 \\
    24 & \textbf{0.558 {\textpm} 0.011} & 0.544 {\textpm} 0.010 & 0.533 {\textpm} 0.015 & 0.521 {\textpm} 0.015 \\
    \bottomrule
    \end{tabular}
\end{table*}

\begin{table*}[htbp]
\centering
\setlength{\tabcolsep}{12pt} 
\renewcommand{\arraystretch}{1.3} 
\caption{Classification results on the fastMRI knee dataset at different undersampling rates. The proposed method achieves comparable results to the baselines, showing a similar drop in performance at undersampling rates of $12\times$ and higher. Best results are highlighted in \textbf{bold}.}
\label{tab:knee_results}
    \begin{tabular}{lcccc}
    \toprule
    Undersampling & kViT (Ours) & ResNet50 & EfficientNet & ViT-Tiny \\
    \midrule
    \multicolumn{5}{l}{AUROC (\textpm{} std)} \\
    0  & 0.845 {\textpm} 0.002 & \textbf{0.890 {\textpm} 0.006} & 0.880 {\textpm} 0.007 & 0.867 {\textpm} 0.007 \\
    2  & 0.839 {\textpm} 0.004 & 0.865 {\textpm} 0.007 & \textbf{0.867 {\textpm} 0.007} & 0.836 {\textpm} 0.010 \\
    4  & 0.834 {\textpm} 0.005 & 0.866 {\textpm} 0.005 & \textbf{0.867 {\textpm} 0.008} & 0.842 {\textpm} 0.012 \\
    6  & 0.832 {\textpm} 0.004 & 0.843 {\textpm} 0.008 & \textbf{0.856 {\textpm} 0.005} & 0.827 {\textpm} 0.017 \\
    8  & 0.830 {\textpm} 0.003 & 0.820 {\textpm} 0.008 & \textbf{0.849 {\textpm} 0.005} & 0.816 {\textpm} 0.015 \\
    10 & 0.822 {\textpm} 0.006 & 0.801 {\textpm} 0.004 & \textbf{0.836 {\textpm} 0.009} & 0.806 {\textpm} 0.013 \\
    12 & 0.737 {\textpm} 0.023 & 0.767 {\textpm} 0.009 & \textbf{0.786 {\textpm} 0.007} & 0.727 {\textpm} 0.019 \\
    16 & 0.682 {\textpm} 0.033 & 0.756 {\textpm} 0.012 & \textbf{0.766 {\textpm} 0.010} & 0.694 {\textpm} 0.019 \\
    24 & 0.668 {\textpm} 0.038 & 0.708 {\textpm} 0.017 & \textbf{0.740 {\textpm} 0.008} & 0.674 {\textpm} 0.008 \\
    \midrule
    \multicolumn{5}{l}{AUPRC (\textpm{} std)} \\
    0  & 0.494 {\textpm} 0.005 & \textbf{0.552 {\textpm} 0.012} & 0.536 {\textpm} 0.013 & 0.520 {\textpm} 0.006 \\
    2  & 0.489 {\textpm} 0.005 & 0.513 {\textpm} 0.012 & \textbf{0.523 {\textpm} 0.016} & 0.492 {\textpm} 0.007 \\
    4  & 0.487 {\textpm} 0.007 & \textbf{0.524 {\textpm} 0.006} & 0.519 {\textpm} 0.014 & 0.499 {\textpm} 0.007 \\
    6  & 0.486 {\textpm} 0.008 & 0.494 {\textpm} 0.012 & \textbf{0.510 {\textpm} 0.011} & 0.484 {\textpm} 0.010 \\
    8  & 0.485 {\textpm} 0.006 & 0.465 {\textpm} 0.012 & \textbf{0.498 {\textpm} 0.014} & 0.474 {\textpm} 0.010 \\
    10 & \textbf{0.483 {\textpm} 0.004} & 0.449 {\textpm} 0.008 & 0.476 {\textpm} 0.014 & 0.461 {\textpm} 0.015 \\
    12 & 0.410 {\textpm} 0.010 & 0.412 {\textpm} 0.011 & \textbf{0.423 {\textpm} 0.017} & 0.385 {\textpm} 0.017 \\
    16 & 0.368 {\textpm} 0.018 & \textbf{0.405 {\textpm} 0.009} & 0.402 {\textpm} 0.016 & 0.354 {\textpm} 0.019 \\
    24 & 0.356 {\textpm} 0.019 & 0.369 {\textpm} 0.008 & \textbf{0.380 {\textpm} 0.014} & 0.340 {\textpm} 0.008 \\
    \bottomrule
    \end{tabular}
\end{table*}

\begin{table*}[htbp]
\centering
\setlength{\tabcolsep}{12pt} 
\renewcommand{\arraystretch}{1.3} 
\caption{Classification results on the MIL dataset at different undersampling rates. While the proposed method achieves similar AUROC results as the baselines, it excels in AUPRC at every tested undersampling rate. Best results are highlighted in \textbf{bold}.}
\label{tab:mil_results}
    \begin{tabular}{lcccc}
    \toprule
    Undersampling & kViT (Ours) & ResNet50 & EfficientNet & ViT-Tiny \\
    \midrule
    \multicolumn{5}{l}{AUROC (\textpm{} std)} \\
    0  & 0.645 {\textpm} 0.083 & 0.643 {\textpm} 0.057 & 0.528 {\textpm} 0.069 & \textbf{0.654 {\textpm} 0.099} \\
    2  & 0.651 {\textpm} 0.080 & 0.638 {\textpm} 0.027 & 0.545 {\textpm} 0.106 & \textbf{0.670 {\textpm} 0.102} \\
    4  & 0.634 {\textpm} 0.094 & 0.629 {\textpm} 0.037 & 0.518 {\textpm} 0.119 & \textbf{0.665 {\textpm} 0.097} \\
    6  & 0.630 {\textpm} 0.071 & 0.621 {\textpm} 0.026 & 0.546 {\textpm} 0.114 & \textbf{0.661 {\textpm} 0.099} \\
    8  & 0.626 {\textpm} 0.071 & 0.619 {\textpm} 0.040 & 0.541 {\textpm} 0.114 & \textbf{0.661 {\textpm} 0.096} \\
    10 & 0.621 {\textpm} 0.067 & 0.618 {\textpm} 0.040 & 0.512 {\textpm} 0.132 & \textbf{0.627 {\textpm} 0.088} \\
    12 & \textbf{0.646 {\textpm} 0.043} & 0.619 {\textpm} 0.014 & 0.544 {\textpm} 0.107 & 0.570 {\textpm} 0.081 \\
    16 & \textbf{0.683 {\textpm} 0.101} & 0.633 {\textpm} 0.019 & 0.550 {\textpm} 0.080 & 0.552 {\textpm} 0.057 \\
    24 & \textbf{0.643 {\textpm} 0.085} & 0.586 {\textpm} 0.049 & 0.588 {\textpm} 0.030 & 0.533 {\textpm} 0.057 \\
    \midrule
    \multicolumn{5}{l}{AUPRC (\textpm{} std)} \\
    0  & \textbf{0.541 {\textpm} 0.091} & 0.530 {\textpm} 0.053 & 0.449 {\textpm} 0.057 & 0.514 {\textpm} 0.082 \\
    2  & \textbf{0.545 {\textpm} 0.069} & 0.514 {\textpm} 0.027 & 0.450 {\textpm} 0.098 & 0.534 {\textpm} 0.083 \\
    4  & \textbf{0.547 {\textpm} 0.063} & 0.506 {\textpm} 0.037 & 0.447 {\textpm} 0.118 & 0.521 {\textpm} 0.073 \\
    6  & \textbf{0.534 {\textpm} 0.049} & 0.503 {\textpm} 0.012 & 0.461 {\textpm} 0.085 & 0.527 {\textpm} 0.089 \\
    8  & \textbf{0.528 {\textpm} 0.062} & 0.497 {\textpm} 0.026 & 0.438 {\textpm} 0.091 & 0.513 {\textpm} 0.078 \\
    10 & \textbf{0.527 {\textpm} 0.054} & 0.490 {\textpm} 0.034 & 0.427 {\textpm} 0.086 & 0.489 {\textpm} 0.076 \\
    12 & \textbf{0.531 {\textpm} 0.029} & 0.476 {\textpm} 0.014 & 0.445 {\textpm} 0.063 & 0.456 {\textpm} 0.084 \\
    16 & \textbf{0.576 {\textpm} 0.071} & 0.488 {\textpm} 0.025 & 0.453 {\textpm} 0.061 & 0.447 {\textpm} 0.056 \\
    24 & \textbf{0.543 {\textpm} 0.050} & 0.450 {\textpm} 0.033 & 0.489 {\textpm} 0.032 & 0.430 {\textpm} 0.062 \\
    \bottomrule
    \end{tabular}
\end{table*}

\begin{figure*}[htbp]
    \centering
    \includegraphics[width=0.8\linewidth]{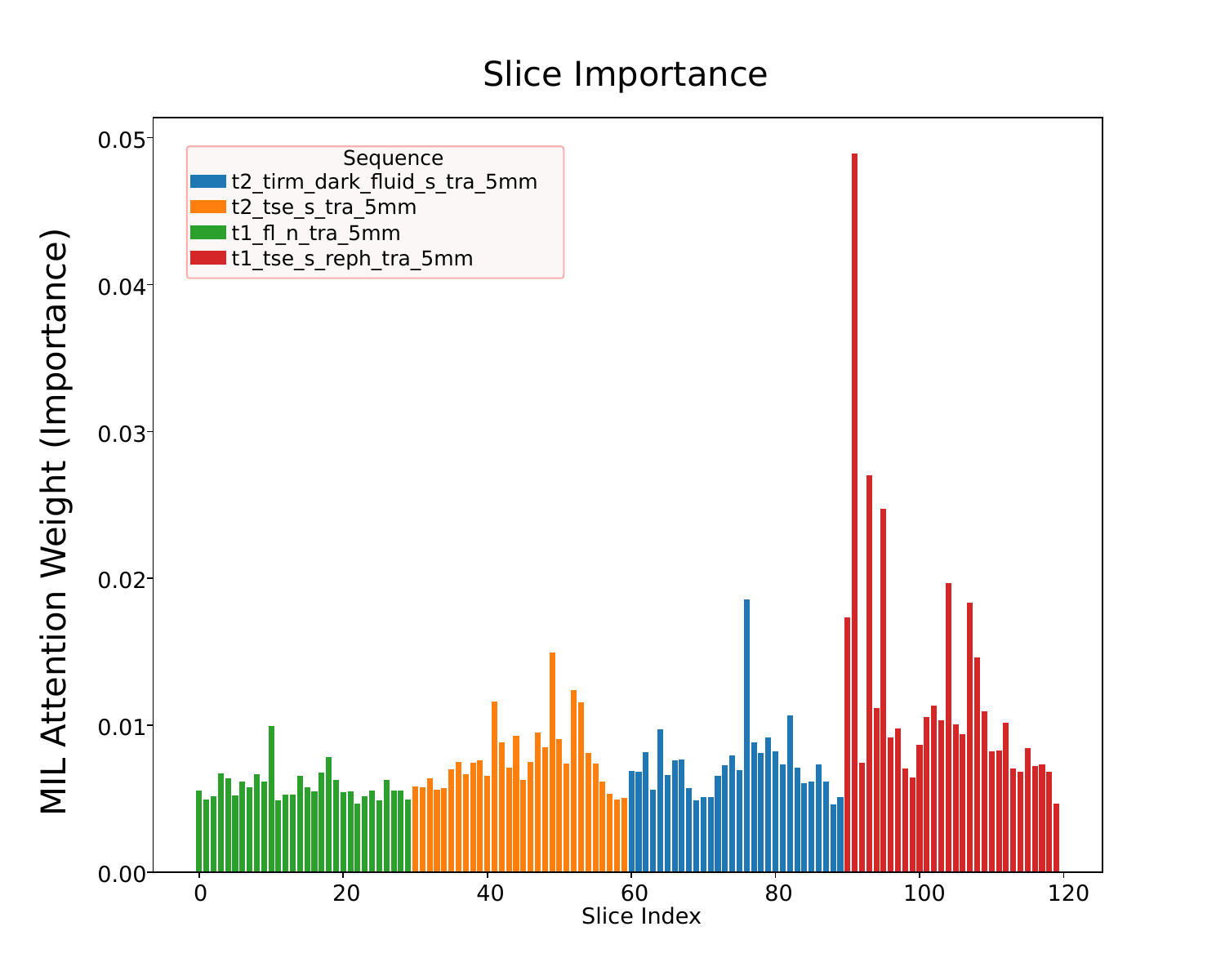}
    \caption{Slice importance scores across multiple sequences of one patient from the in-house MIL dataset.}
    \label{fig:mil_slice_importance}
\end{figure*}

\begin{figure*}[htbp]
    \centering
    \includegraphics[width=0.5\linewidth]{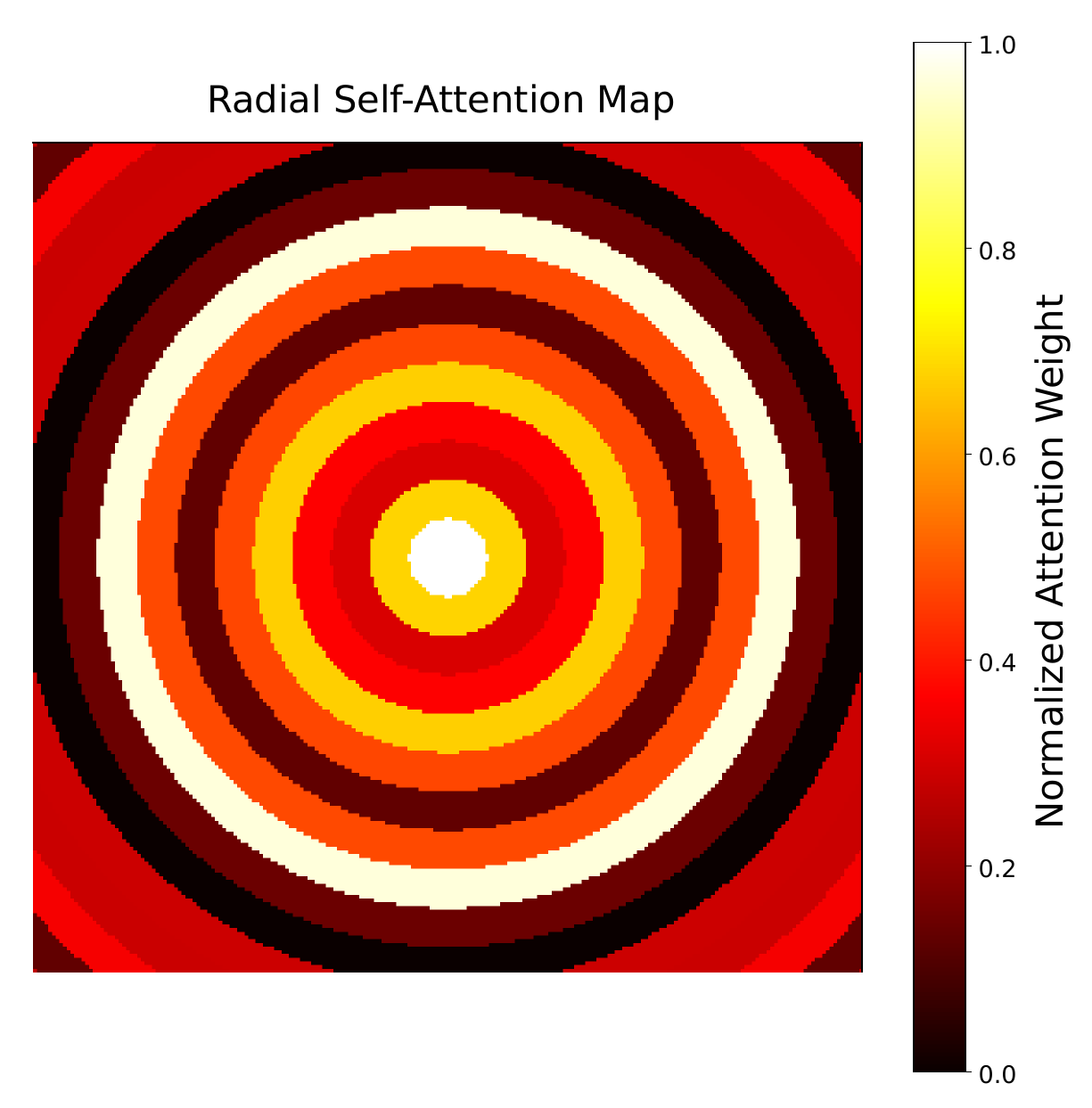}
    \caption{Attention map from the last Transformer layer of the proposed method on the most relevant slice in the MIL setting on the in-house dataset.}
    \label{fig:mil_attn_map}
\end{figure*}

%% file: main.bib
@article{eggerMedicalDeepLearning2022b,
  title = {Medical Deep Learning—{{A}} Systematic Meta-Review},
  author = {Egger, Jan and Gsaxner, Christina and Pepe, Antonio and Pomykala, Kelsey L and Jonske, Frederic and Kurz, Manuel and Li, Jianning and Kleesiek, Jens},
  date = {2022},
  journaltitle = {Computer methods and programs in biomedicine},
  shortjournal = {Computer methods and programs in biomedicine},
  volume = {221},
  pages = {106874},
  publisher = {Elsevier},
  issn = {0169-2607}
}

@article{bouhafraDeepLearningApproaches2025,
  title = {Deep Learning Approaches for Brain Tumor Detection and Classification Using {{MRI}} Images (2020 to 2024): A Systematic Review},
  author = {Bouhafra, Sara and El Bahi, Hassan},
  date = {2025},
  journaltitle = {Journal of Imaging Informatics in Medicine},
  shortjournal = {Journal of Imaging Informatics in Medicine},
  volume = {38},
  number = {3},
  pages = {1403--1433},
  publisher = {Springer},
  issn = {2948-2933}
}

@article{heckelDeepLearningAccelerated2024a,
  title = {Deep Learning for Accelerated and Robust {{MRI}} Reconstruction},
  author = {Heckel, Reinhard and Jacob, Mathews and Chaudhari, Akshay and Perlman, Or and Shimron, Efrat},
  date = {2024},
  journaltitle = {Magnetic Resonance Materials in Physics, Biology and Medicine},
  shortjournal = {Magnetic Resonance Materials in Physics, Biology and Medicine},
  volume = {37},
  number = {3},
  pages = {335--368},
  publisher = {Springer},
  issn = {1352-8661}
}

@article{rempeTumorLikelihoodEstimation2024b,
  title = {Tumor Likelihood Estimation on {{MRI}} Prostate Data by Utilizing K-{{Space}} Information},
  author = {Rempe, Moritz and Hörst, Fabian and Seibold, Constantin and Hadaschik, Boris and Schlimbach, Marco and Egger, Jan and Kröninger, Kevin and Breuer, Felix and Blaimer, Martin and Kleesiek, Jens},
  date = {2024},
  journaltitle = {arXiv preprint arXiv:2407.06165},
  eprint = {2407.06165},
  eprinttype = {arXiv}
}

@inproceedings{liClassificationRegressionSegmentation2024a,
  title = {Classification, {{Regression}} and {{Segmentation}} Directly from k-{{Space}} in {{Cardiac MRI}}},
  booktitle = {International {{Workshop}} on {{Machine Learning}} in {{Medical Imaging}}},
  author = {Li, Ruochen and Pan, Jiazhen and Zhu, Youxiang and Ni, Juncheng and Rueckert, Daniel},
  year = 2024,
  pages = {31--41},
  publisher = {Springer}
}

@article{douMRIDenoisingNonblind2025,
  title = {{{MRI}} Denoising with a Non‐blind Deep Complex‐valued Convolutional Neural Network},
  author = {Dou, Quan and Wang, Zhixing and Feng, Xue and Campbell‐Washburn, Adrienne E and Mugler III, John P and Meyer, Craig H},
  date = {2025},
  journaltitle = {NMR in Biomedicine},
  shortjournal = {NMR in Biomedicine},
  volume = {38},
  number = {1},
  pages = {e5291},
  publisher = {Wiley Online Library},
  issn = {0952-3480}
}

@article{baiTOPSspeedComplexvaluedConvolutional2025,
  title = {{{TOPS-speed}} Complex-Valued Convolutional Accelerator for Feature Extraction and Inference},
  author = {Bai, Yunping and Xu, Yifu and Chen, Shifan and Zhu, Xiaotian and Wang, Shuai and Huang, Sirui and Song, Yuhang and Zheng, Yixuan and Liu, Zhihui and Tan, Sim},
  date = {2025},
  journaltitle = {Nature Communications},
  shortjournal = {Nature Communications},
  volume = {16},
  number = {1},
  pages = {292},
  publisher = {Nature Publishing Group UK London},
  issn = {2041-1723}
}

@article{yuComplexvaluedNeuralnetworkbasedFederated2024,
  title = {Complex-Valued Neural-Network-Based Federated Learning for Multiuser Indoor Positioning Performance Optimization},
  author = {Yu, Hanzhi and Liu, Yuchen and Chen, Mingzhe},
  date = {2024},
  journaltitle = {IEEE Internet of Things Journal},
  shortjournal = {IEEE Internet of Things Journal},
  volume = {11},
  number = {21},
  pages = {34065--34077},
  publisher = {IEEE},
  issn = {2327-4662}
}

@article{zhaoReviewConvolutionalNeural2024,
  title = {A Review of Convolutional Neural Networks in Computer Vision},
  author = {Zhao, Xia and Wang, Limin and Zhang, Yufei and Han, Xuming and Deveci, Muhammet and Parmar, Milan},
  date = {2024},
  journaltitle = {Artificial Intelligence Review},
  shortjournal = {Artificial Intelligence Review},
  volume = {57},
  number = {4},
  pages = {99},
  publisher = {Springer},
  issn = {1573-7462}
}

@article{takahashiComparisonVisionTransformers2024,
  title = {Comparison of Vision Transformers and Convolutional Neural Networks in Medical Image Analysis: {{A}} Systematic Review},
  author = {Takahashi, Satoshi and Sakaguchi, Yusuke and Kouno, Nobuji and Takasawa, Ken and Ishizu, Kenichi and Akagi, Yu and Aoyama, Rina and Teraya, Naoki and Bolatkan, Amina and Shinkai, Norio},
  date = {2024},
  journaltitle = {Journal of Medical Systems},
  shortjournal = {Journal of Medical Systems},
  volume = {48},
  number = {1},
  pages = {84},
  publisher = {Springer},
  issn = {1573-689X}
}

@article{dosovitskiyImageWorth16x162020,
  title = {An Image Is Worth 16x16 Words: {{Transformers}} for Image Recognition at Scale},
  author = {Dosovitskiy, Alexey},
  date = {2020},
  journaltitle = {arXiv preprint arXiv:2010.11929},
  eprint = {2010.11929},
  eprinttype = {arXiv}
}

@inproceedings{liuSwinTransformerHierarchical2021,
  title = {Swin Transformer: {{Hierarchical}} Vision Transformer Using Shifted Windows},
  author = {Liu, Ze and Lin, Yutong and Cao, Yue and Hu, Han and Wei, Yixuan and Zhang, Zheng and Lin, Stephen and Guo, Baining},
  date = {2021},
  pages = {10012--10022},
  eventtitle = {Proceedings of the {{IEEE}}/{{CVF}} International Conference on Computer Vision}
}

@inproceedings{chenDptDeformablePatchbased2021,
  title = {Dpt: {{Deformable}} Patch-Based Transformer for Visual Recognition},
  author = {Chen, Zhiyang and Zhu, Yousong and Zhao, Chaoyang and Hu, Guosheng and Zeng, Wei and Wang, Jinqiao and Tang, Ming},
  date = {2021},
  pages = {2899--2907},
  eventtitle = {Proceedings of the 29th {{ACM}} International Conference on Multimedia}
}

@inproceedings{athwaleDarswinDistortionAware2023,
  title = {Darswin: {{Distortion}} Aware Radial Swin Transformer},
  author = {Athwale, Akshaya and Afrasiyabi, Arman and Lagüe, Justin and Shili, Ichrak and Ahmad, Ola and Lalonde, Jean-François},
  date = {2023},
  pages = {5929--5938},
  eventtitle = {Proceedings of the {{IEEE}}/{{CVF International Conference}} on {{Computer Vision}}}
}

@article{alkhatibPolSARImageClassification2025,
  title = {{{PolSAR}} Image Classification Using Complex-Valued Multiscale Attention Vision Transformer ({{CV-MsAtViT}})},
  author = {Alkhatib, Mohammed Q},
  date = {2025},
  journaltitle = {International Journal of Applied Earth Observation and Geoinformation},
  shortjournal = {International Journal of Applied Earth Observation and Geoinformation},
  volume = {137},
  pages = {104412},
  publisher = {Elsevier},
  issn = {1569-8432}
}

@article{rempeKstripNovelSegmentation2024c,
  title = {K-Strip: {{A}} Novel Segmentation Algorithm in k-Space for the Application of Skull Stripping},
  author = {Rempe, Moritz and Mentzel, Florian and Pomykala, Kelsey L and Haubold, Johannes and Nensa, Felix and Kroeninger, Kevin and Egger, Jan and Kleesiek, Jens},
  date = {2024},
  journaltitle = {Computer Methods and Programs in Biomedicine},
  shortjournal = {Computer Methods and Programs in Biomedicine},
  volume = {243},
  pages = {107912},
  publisher = {Elsevier},
  issn = {0169-2607}
}

@article{yenAdaptiveSamplingKspace2024a,
  title = {Adaptive Sampling of K-Space in Magnetic Resonance for Rapid Pathology Prediction},
  author = {Yen, Chen-Yu and Singhal, Raghav and Sharma, Umang and Ranganath, Rajesh and Chopra, Sumit and Pinto, Lerrel},
  date = {2024},
  journaltitle = {arXiv preprint arXiv:2406.04318},
  eprint = {2406.04318},
  eprinttype = {arXiv}
}

@article{manzariMedViTRobustVision2023,
  title = {{{MedViT}}: A Robust Vision Transformer for Generalized Medical Image Classification},
  author = {Manzari, Omid Nejati and Ahmadabadi, Hamid and Kashiani, Hossein and Shokouhi, Shahriar B and Ayatollahi, Ahmad},
  date = {2023},
  journaltitle = {Computers in biology and medicine},
  shortjournal = {Computers in biology and medicine},
  volume = {157},
  pages = {106791},
  publisher = {Elsevier},
  issn = {0010-4825}
}

@article{joyViTADLeveragingModified2025,
  title = {{{ViTAD}}: {{Leveraging}} Modified Vision Transformer for {{Alzheimer}}’s Disease Multi-Stage Classification from Brain {{MRI}} Scans},
  author = {Joy, Md Ashif Mahmud and Nasrin, Shamima and Siddiqua, Ayesha and Farid, Dewan Md},
  date = {2025},
  journaltitle = {Brain Research},
  shortjournal = {Brain Research},
  volume = {1847},
  pages = {149302},
  publisher = {Elsevier},
  issn = {0006-8993}
}

@article{liComplexvaluedTransformerAutomatic2024,
  title = {A Complex-Valued Transformer for Automatic Modulation Recognition},
  author = {Li, Weihao and Deng, Wen and Wang, Keren and You, Ling and Huang, Zhitao},
  date = {2024},
  journaltitle = {IEEE Internet of Things Journal},
  shortjournal = {IEEE Internet of Things Journal},
  volume = {11},
  number = {12},
  pages = {22197--22207},
  publisher = {IEEE},
  issn = {2327-4662}
}

@article{rempeTumorLikelihoodEstimation2024c,
  title = {Tumor Likelihood Estimation on {{MRI}} Prostate Data by Utilizing K-{{Space}} Information},
  author = {Rempe, Moritz and Hörst, Fabian and Seibold, Constantin and Hadaschik, Boris and Schlimbach, Marco and Egger, Jan and Kröninger, Kevin and Breuer, Felix and Blaimer, Martin and Kleesiek, Jens},
  date = {2024},
  journaltitle = {arXiv preprint arXiv:2407.06165},
  eprint = {2407.06165},
  eprinttype = {arXiv}
}

@article{suRoformerEnhancedTransformer2024,
  title = {Roformer: {{Enhanced}} Transformer with Rotary Position Embedding},
  author = {Su, Jianlin and Ahmed, Murtadha and Lu, Yu and Pan, Shengfeng and Bo, Wen and Liu, Yunfeng},
  date = {2024},
  journaltitle = {Neurocomputing},
  shortjournal = {Neurocomputing},
  volume = {568},
  pages = {127063},
  publisher = {Elsevier},
  issn = {0925-2312}
}

@inproceedings{heoRotaryPositionEmbedding2024,
  title = {Rotary Position Embedding for Vision Transformer},
  author = {Heo, Byeongho and Park, Song and Han, Dongyoon and Yun, Sangdoo},
  date = {2024},
  pages = {289--305},
  publisher = {Springer},
  eventtitle = {European {{Conference}} on {{Computer Vision}}}
}

@article{tibrewalaFastMRIProstatePublic2024b,
  title = {{{FastMRI Prostate}}: {{A}} Public, Biparametric {{MRI}} Dataset to Advance Machine Learning for Prostate Cancer Imaging},
  author = {Tibrewala, Radhika and Dutt, Tarun and Tong, Angela and Ginocchio, Luke and Lattanzi, Riccardo and Keerthivasan, Mahesh B and Baete, Steven H and Chopra, Sumit and Lui, Yvonne W and Sodickson, Daniel K},
  date = {2024},
  journaltitle = {Scientific data},
  shortjournal = {Scientific data},
  volume = {11},
  number = {1},
  pages = {404},
  publisher = {Nature Publishing Group UK London},
  issn = {2052-4463}
}

@article{knollFastMRIPubliclyAvailable2020a,
  title = {{{fastMRI}}: {{A}} Publicly Available Raw k-Space and {{DICOM}} Dataset of Knee Images for Accelerated {{MR}} Image Reconstruction Using Machine Learning},
  author = {Knoll, Florian and Zbontar, Jure and Sriram, Anuroop and Muckley, Matthew J and Bruno, Mary and Defazio, Aaron and Parente, Marc and Geras, Krzysztof J and Katsnelson, Joe and Chandarana, Hersh},
  date = {2020},
  journaltitle = {Radiology: Artificial Intelligence},
  shortjournal = {Radiology: Artificial Intelligence},
  volume = {2},
  number = {1},
  pages = {e190007},
  publisher = {Radiological Society of North America},
  issn = {2638-6100}
}

@article{zhaoFastMRIClinicalPathology2022,
  title = {{{fastMRI}}+, Clinical Pathology Annotations for Knee and Brain Fully Sampled Magnetic Resonance Imaging Data},
  author = {Zhao, Ruiyang and Yaman, Burhaneddin and Zhang, Yuxin and Stewart, Russell and Dixon, Austin and Knoll, Florian and Huang, Zhengnan and Lui, Yvonne W and Hansen, Michael S and Lungren, Matthew P},
  date = {2022},
  journaltitle = {Scientific Data},
  shortjournal = {Scientific Data},
  volume = {9},
  number = {1},
  pages = {152},
  publisher = {Nature Publishing Group UK London},
  issn = {2052-4463}
}

@article{turkbeyProstateImagingReporting2019b,
  title = {Prostate Imaging Reporting and Data System Version 2.1: 2019 Update of Prostate Imaging Reporting and Data System Version 2},
  author = {Turkbey, Baris and Rosenkrantz, Andrew B and Haider, Masoom A and Padhani, Anwar R and Villeirs, Geert and Macura, Katarzyna J and Tempany, Clare M and Choyke, Peter L and Cornud, Francois and Margolis, Daniel J},
  date = {2019},
  journaltitle = {European urology},
  shortjournal = {European urology},
  volume = {76},
  number = {3},
  pages = {340--351},
  publisher = {Elsevier},
  issn = {0302-2838}
}

@inproceedings{tanEfficientnetRethinkingModel2019,
  title = {Efficientnet: {{Rethinking}} Model Scaling for Convolutional Neural Networks},
  author = {Tan, Mingxing and Le, Quoc},
  date = {2019},
  pages = {6105--6114},
  publisher = {PMLR},
  eventtitle = {International Conference on Machine Learning},
  isbn = {2640-3498}
}

@inproceedings{heDeepResidualLearning2016a,
  title = {Deep Residual Learning for Image Recognition},
  author = {He, Kaiming and Zhang, Xiangyu and Ren, Shaoqing and Sun, Jian},
  date = {2016},
  pages = {770--778},
  eventtitle = {Proceedings of the {{IEEE}} Conference on Computer Vision and Pattern Recognition}
}

@article{dietterichSolvingMultipleInstance1997,
  title = {Solving the Multiple Instance Problem with Axis-Parallel Rectangles},
  author = {Dietterich, Thomas G and Lathrop, Richard H and Lozano-Pérez, Tomás},
  date = {1997},
  journaltitle = {Artificial intelligence},
  shortjournal = {Artificial intelligence},
  volume = {89},
  number = {1--2},
  pages = {31--71},
  publisher = {Elsevier},
  issn = {0004-3702}
}

@article{loshchilovDecoupledWeightDecay2017a,
  title = {Decoupled Weight Decay Regularization},
  author = {Loshchilov, Ilya and Hutter, Frank},
  date = {2017},
  journaltitle = {arXiv preprint arXiv:1711.05101},
  eprint = {1711.05101},
  eprinttype = {arXiv}
}

@article{wibmerHaralickTextureAnalysis2015,
  title = {Haralick Texture Analysis of Prostate {{MRI}}: Utility for Differentiating Non-Cancerous Prostate from Prostate Cancer and Differentiating Prostate Cancers with Different {{Gleason}} Scores},
  author = {Wibmer, Andreas and Hricak, Hedvig and Gondo, Tatsuo and Matsumoto, Kazuhiro and Veeraraghavan, Harini and Fehr, Duc and Zheng, Junting and Goldman, Debra and Moskowitz, Chaya and Fine, Samson W},
  date = {2015},
  journaltitle = {European radiology},
  shortjournal = {European radiology},
  volume = {25},
  number = {10},
  pages = {2840--2850},
  publisher = {Springer},
  issn = {0938-7994}
}

@article{desmetUseTwoslicetouchRule2006,
  title = {Use of the “Two-Slice-Touch” Rule for the {{MRI}} Diagnosis of Meniscal Tears},
  author = {De Smet, Arthur A and Tuite, Michael J},
  date = {2006},
  journaltitle = {American Journal of Roentgenology},
  shortjournal = {American Journal of Roentgenology},
  volume = {187},
  number = {4},
  pages = {911--914},
  publisher = {American Roentgen Ray Society},
  issn = {0361-803X}
}

@article{rempePhaseGenDiffusionBasedApproach2025a,
  title = {{{PhaseGen}}: {{A Diffusion-Based Approach}} for {{Complex-Valued MRI Data Generation}}},
  author = {Rempe, Moritz and Hörst, Fabian and Becker, Helmut and Schlimbach, Marco and Rotkopf, Lukas and Kröninger, Kevin and Kleesiek, Jens},
  date = {2025},
  journaltitle = {arXiv preprint arXiv:2504.07560},
  eprint = {2504.07560},
  eprinttype = {arXiv}
}

@misc{rw2019timm,
  author = {Ross Wightman},
  title = {PyTorch Image Models},
  year = {2019},
  publisher = {GitHub},
  journal = {GitHub repository},
  doi = {10.5281/zenodo.4414861},
  howpublished = {\url{https://github.com/rwightman/pytorch-image-models}}
}
